\documentclass{article}
\usepackage{spconf,amsmath,amssymb,graphicx,hyperref}
\usepackage{booktabs,cleveref,subcaption,comment,float,multirow}
\usepackage[numbers]{natbib}
\usepackage[table]{xcolor}

\title{Learnable Instance Attention Filtering for Adaptive Detector Distillation}
\name{%
  Chen Liu$^{1*}$,
  Qizhen Lan$^{1*}$,
  Zhicheng Ding$^{2}$,
  Xinyu Chu$^{2}$,
  Qing Tian$^{1\dagger}$
  \thanks{$*$ These two student authors contributed equally.}%
  \thanks{$\dagger$ Corresponding author. This work was supported by the National Science Foundation under Award Nos. 2153404 and 2412285.}%
}
\address{%
  $^{1}$Dept. of CS, University of Alabama at Birmingham \quad
  $^{2}$Dept. of CS, Bowling Green State University%
}

\begin{document}
%
\maketitle
\begin{abstract}
As deep vision models grow increasingly complex to achieve higher performance, deployment efficiency has become a critical concern. Knowledge distillation (KD) mitigates this issue by transferring knowledge from large teacher models to compact student models. While many feature-based KD methods rely on spatial filtering to guide distillation, they typically treat all object instances uniformly, ignoring instance-level variability. Moreover, existing attention filtering mechanisms are typically heuristic or teacher-driven, rather than learned with the student. To address these limitations, we propose Learnable Instance Attention Filtering for Adaptive Detector Distillation (LIAF-KD), a novel framework that introduces learnable instance selectors to dynamically evaluate and reweight instance importance during distillation. Notably, the student contributes to this process based on its evolving learning state. Experiments on the KITTI and COCO datasets demonstrate consistent improvements, with a 2\% gain on a GFL ResNet-50 student without added complexity, outperforming state-of-the-art methods.
\end{abstract}
\section{Introduction}
\label{sec:intro}

As deep learning models \cite{yang2026tooltree,du2026unsupervised,10.1007/978-3-031-72970-6_10,Su_2026_WACV,ma2024generative,ma2025ms,ma2026gor} continue to grow in complexity to achieve higher performance, their deployment on edge devices becomes increasingly impractical~\cite{zhang2023tdec,zheng2024deep,qu2024conditional,qu2025end,ZHANG2026133318,sun2026align,li2025pointdico,yang2026evotool}. Knowledge Distillation (KD) \cite{hinton2015distilling} addresses this by transferring knowledge from one or more large teacher models to a compact student. While effective in image classification, KD remains relatively underexplored for dense prediction tasks.

Instead of logit distillation (as in vanilla KD \cite{hinton2015distilling}), feature distillation, pioneered by FitNet \cite{romero2014fitnets}, is more suitable for detection tasks as it transfers rich spatial and structural information from intermediate layers. Since not all pixels contribute equally to the effectiveness of distillation, a number of studies (e.g., \cite{Wang_2019_CVPR, Dai_2021_CVPR, lan2022adaptive}) have proposed spatially-aware distillation techniques to selectively emphasize informative locations. However, many rely on man-made heuristics (e.g., activation strength, anchor locations, and handcrafted saliency). Huang et al.~\cite{huang2023masked} propose masked distillation using receptive tokens to improve efficiency and flexibility; however, it relies only on the teacher model to determine where to focus during distillation and may suffer from tokens capturing irrelevant background noise. As widely acknowledged in the detection distillation literature \cite{Wang_2019_CVPR, Guo_2021_CVPR, menteidis2024foreground}, foreground pixels play a more critical role and thus should carry more weight. However, naively emphasizing foreground regions overlooks instance variability. The above gaps motivate our instance-adaptive distillation strategy, which dynamically assesses and weighs each instance with active student involvement, adapting to the student’s evolving needs.

In this paper, we propose Learnable Instance Attention Filtering for Adaptive Detector Distillation (LIAF-KD), an instance-adaptive and student-aware method. Specifically, prior to distillation, we extract ROI-aligned instance features from the teacher and train a set of learnable instance selectors that can evaluate an instance's importance based on its appearance. During distillation, these selectors generate scores based on both the teacher's knowledge and the student's evolving learning state. These scores are then used to weight the features during distillation, ensuring that more informative instances contribute more strongly to knowledge transfer.
Unlike prior teacher-centric, instance-agnostic approaches (e.g., \cite{huang2023masked}), our method involves active student participation in selecting which instances to pay more attention to during distillation.


\section{Related Work}
\label{sec:relatedwork}

\textbf{Object Detection} is a fundamental task in computer vision that involves simultaneous object localization and classification. Recent advances have shifted from traditional handcrafted pipelines to deep learning architectures \cite{qu2025robust}, which achieve remarkable accuracy.
Detectors based on deep convolutional networks can be categorized into two-stage \cite{girshick2015fast,Ren_2017,ma2025fine} and one-stage \cite{Liu_2016,redmon2016you,lin2017focal} detectors. Two-stage detectors generally achieve high accuracy by generating region proposals and refining predictions at the cost of slower inference speed. In contrast, one-stage detectors eliminate the proposal stage and offer faster inference by making dense predictions in a single pass. Thus, one-stage detectors (e.g., \cite{redmon2016you,lin2017focal,li2020generalized}) are more desirable in real-world edge scenarios due to their relatively high efficiency. 
More recently, transformer-based detectors~\cite{detr, meng2021-CondDETR,zhang2022dino} have gained traction.
However, the absence of convolutional inductive biases results in significantly higher computational costs. Given our goal of improving efficiency, this paper focuses on convolutional architectures.

\textbf{Knowledge distillation (KD)} trains a smaller student model to mimic a larger teacher model~\cite{hinton2015distilling}. While KD has been extensively studied in image classification~\cite{gou2021knowledge}, its application to dense prediction tasks remains challenging, largely due to their multi-task nature and the overwhelming dominance of background pixels. Existing KD approaches for object detection are predominantly feature-based, as feature distillation provides more localized supervision than logit distillation. However, most existing methods lack instance adaptivity and/or rely on heuristic designs~\cite{Wang_2019_CVPR,Guo_2021_CVPR,Dai_2021_CVPR}.
Although still instance-nonadaptive, MasKD~\cite{huang2023masked} takes a step toward learned region selection by generating spatial masks using receptive tokens trained by the teacher. However, these masks remain fixed during distillation and do not reflect the student’s learning dynamics. Building on this idea, FreeKD~\cite{Zhang_2024_CVPR} shifts knowledge distillation to the frequency domain by introducing semantic frequency prompts that generate teacher-driven attention masks. Yet, as in MasKD, the selection of focus regions occurs without active student involvement, with a uniform teacher-determined strategy applied across instances.


\section{Methodology}\label{sec:methodology}

\subsection{Learnable Instance Attention Filtering for Adaptive Detector Distillation}

\begin{figure*}[h]
    \centering
    \includegraphics[width=0.685\textwidth,keepaspectratio]{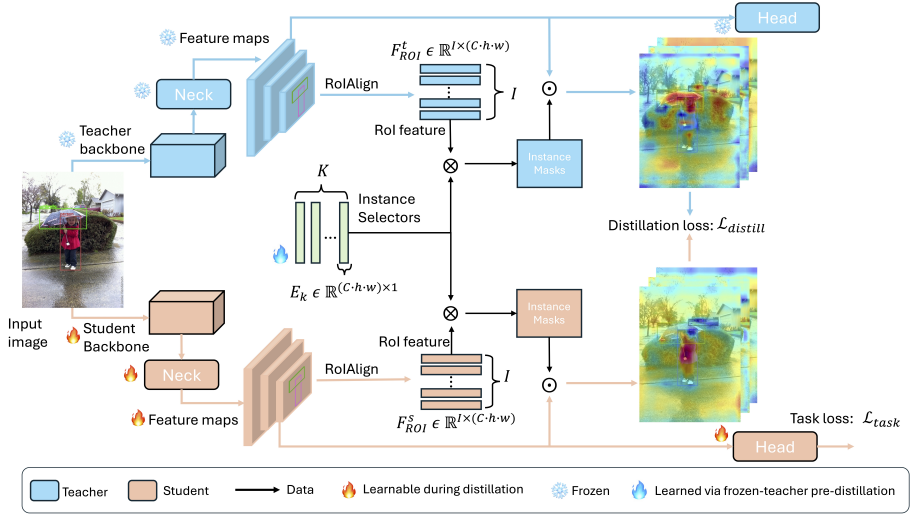}
    \caption{Overview of the proposed LIAF-KD framework. It employs instance selectors to dynamically reweight instances during distillation, guided by both the teacher’s knowledge and the student’s learning dynamics.}
    \label{fig:overview}
\end{figure*}

In knowledge distillation for dense prediction tasks, not all pixels contribute equally. A common strategy is to apply a spatial mask $\mathbf{M}$ that selectively highlights meaningful regions. This process can be expressed as follows:
\begin{equation}
\mathcal{L}_{KD}=\frac{1}{CHW}
\left\lVert \mathbf{M} \odot \left( F^T - f_{\mathrm{\phi}}(F^S) \right) \right\rVert_2^2\ \,.
\end{equation}
\noindent where $F$ represents feature maps, and the superscripts \(T\) and \(S\) represent the teacher and student, respectively. The function \(f_{\phi}\) is a projection layer that aligns \(F^S\) with the dimension of \(F^T\). \(C\), \(H\), and \(W\) respectively denote the number of channels, height, and width of features.
Different feature distillation methods employ different masking strategies ($\mathbf{M}$) to emphasize informative regions.
However, most existing strategies suffer from two limitations: (1) the masks are either heuristic-based or dictated by the teacher without active student involvement; and (2) the same masking strategy is applied uniformly across all instances, ignoring their variability. To address these issues, this paper proposes \textbf{Learnable Instance Attention Filtering for Adaptive Detector Distillation (LIAF-KD)}, a novel framework that leverages learnable instance-aware selectors to dynamically evaluate and reweigh the importance of instances guided by both the teacher's knowledge and student learning dynamics. 
The proposed method proceeds in two stages: instance selector learning (Sec.~\ref{sec:instance_selector_learning}) and student-aware distillation guided by the learned selectors (Sec.~\ref{sec:instance_selector_distill}). Figure~\ref{fig:overview} provides an overview of our framework.

\subsubsection{Instance selector learning}\label{sec:instance_selector_learning}
In this stage, we train an ensemble of $K$ instance selectors \( E_{k} \in \mathbb{R}^{(C \cdot h \cdot w) \times 1} \) using the teacher model. Given a feature block $F$ and the bounding box $B_{ins}$ of instance $ins$, we apply the RoIAlign function \cite{he2017mask} to extract instance-specific features \( F_{ins} \in \mathbb{R}^{C \times h \times w} \), which are then flattened and batched to \(F_{ROI} \in \mathbb{R}^{I \times (C \cdot h \cdot w)}\):
\begin{equation}
    F_{ins} = f_{\mathrm{roi\_align}}(F, B_{ins}) \,,F_{ROI} = f_{\mathrm{batch}}(f_{\mathrm{flatten}} (F_{ins}))\,.
\end{equation}
To achieve improved detection performance, we use learnable instance selectors to emphasize informative features:
\begin{equation}
    A_k = \sigma(F_{ROI} \cdot E_{k}) \in \mathbb{R}^{I \times 1} \,,
    \label{eq:ins_eval}
\end{equation}
\noindent where $\sigma$ stands for the softmax function. After instance selector learning with the teacher's objective $\mathcal{L}_{task}^{T}$, \(A_k\) encodes the instance importance scores given by the selector $k$, and the instance selectors $E$ acquire the ability to evaluate the importance of each instance based on their feature representation.
In addition, we employ the following diversity loss to encourage expertise diversity among the instance selectors:
\begin{equation}
\mathcal{L}_{div} = \frac{2 \sum_{i=1}^{K} \sum_{\substack{j=1 \\ j \ne i}}^{K} E_i \cdot E_j}{\sum_{i=1}^{K} E_i^2 + \sum_{j=1}^{K} E_j^2} \,.
\end{equation}

To summarize, the training loss of pre-distillation instance selector learning is defined as:
\begin{equation}
    \mathcal{L}_{selector}=\mathcal{L}_{task}^{T} (A) +\mu \mathcal{L}_{div}(E) \,,
\end{equation}
\noindent where $\mathcal{L}_{task}^{T}$ denotes the teacher's task loss, and $A$ indicates the incorporation of the instance evaluation process described in Eq.~\ref{eq:ins_eval}. The scalar $\mu$ is a balancing hyperparameter.

\subsubsection{Student-aware distillation with instance selectors}\label{sec:instance_selector_distill}

In the distillation stage, we utilize the learned instance selectors to compute the instance scores \(A\) based on both the teacher's knowledge and the student's evolving features. 
For both the teacher and student models, the attention values are averaged over the \(K\) selectors to produce a weight for each instance:
\begin{equation}
    A^T = \frac{1}{K} \sum_{k=1}^{K} \sigma(F^T_{ROI} \cdot E_{k}) \,,
    A^S = \frac{1}{K} \sum_{k=1}^{K} \sigma(F^S_{ROI} \cdot E_{k}) \,.
    \label{eq:ins_distil}
\end{equation}


These instance scores are then used to construct a spatially aware soft mask. We first initialize a mask \(\mathbf{M} \in \mathbb{R}^{N \times 1 \times H \times W}\) with all values set to 1, where \(N\) represents the batch size.
For each instance \(i = 1, \dots, I\) with attention value \(A_i\) and ROI region
$\mathcal{R}_i = \{(h, w) \mid y_{1i} \leq h < y_{2i},\ x_{1i} \leq w < x_{2i} \}$, the corresponding pixels of \(\mathbf{M}\) are updated as:
\begin{equation}
\mathbf{M}_{b_i, 0, h, w} \leftarrow \mathbf{M}_{b_i, 0, h, w} \cdot A_i, \quad \forall (h, w) \in \mathcal{R}_i \,,
\end{equation}
\noindent where \(b_i\) is the index indicating the image within the batch that contains instance \(i\), and the subscript \(0\) denotes the single channel dimension of the mask.
The mask is then broadcast across the channel dimension to form \(\mathbf{\hat{M}} \in \mathbb{R}^{N \times C \times H \times W}\), and is applied to the feature map (teacher or student) by element-wise multiplication:
\begin{equation}
\hat{F} = F \odot \mathbf{\hat{M}} \,.
\end{equation}
\noindent The overall distillation loss \(\mathcal{L}_{dist}\) can then be defined as:

\begin{equation}
\mathcal{L}_{dist}=\frac{1}{CHW}\left\lVert \hat{F}^{T} - f_{\mathrm{\phi}}(\hat{F}^{S}) \right\rVert_2^2\ \,,
\end{equation}

\noindent where \(\hat{F}^{T}\) and \(\hat{F}^{S}\) represent the instance-reweighted feature maps of the teacher and student models, respectively.

\section{Experiments and Results}\label{sec:experiment}

We demonstrate the effectiveness of our LIAF-KD method on the widely used KITTI \cite{geiger2012we} and MS COCO \cite{lin2014microsoft} datasets. Following \cite{lan2024gradient}, similar categories are merged for KITTI.
Distillation is performed at the FPN neck \cite{lin2017feature} and evaluated using GFL \cite{li2020generalized} and RetinaNet \cite{lin2017focal} detectors, with ResNet-101 and ResNet-50 serving as the teacher and student backbones, respectively. 
All models are optimized with SGD using a momentum of 0.9 and a weight decay of 0.0001. The number of instance selectors \(\textit{K}\) is set to 6.






\subsection{Quantitative Results}

\textbf{Results on KITTI.} 
According to Table \ref{tab:kitti-gfl and retinanet}, for both the GFL and RetinaNet detectors, LIAF-KD outperforms the student baseline and other KD competitors by clear margins. In the GFL case, it even outperforms the teacher model by 1.4\%.
Our LIAF-KD outperforms the MasKD by clear margins even without inheriting its pixel-wise selection strategy.
Additionally, the hybrid LIAF-KD+MasKD approach achieves the best performance. That said, the performance gain from introducing MasKD into our LIAF-KD is limited, demonstrating the independence and effectiveness of our instance-adaptive and student-aware KD method.


\begin{table}[!h]
  \centering
  \resizebox{0.48\textwidth}{!}{
  \begin{tabular}{@{}lcccccccc@{}}
    \specialrule{1pt}{0pt}{0pt}
    Method & mAP & $AP_{50}$ & $AP_{75}$ & $AP_{S}$ & $AP_{M}$ & $AP_{L}$ \\\hline
    T: GFL-R101   & 61.3 & 87.6 & 68.8 & 52.8 & 61.6 & 67.3 \\
    S: GFL-R50    & 59.5 & 86.9 & 66.2 & 50.8 & 59.6 & 65.2 \\\hline
    DeFeat \cite{Guo_2021_CVPR}     & 61.3 & 88.2 & 68.8 & 53.5 & 61.4 & 66.4 \\
    FGFI \cite{Wang_2019_CVPR}      & 61.4 & 88.4 & 69.9 & 53.6 & 61.1 & 68.0 \\
    GID \cite{Dai_2021_CVPR}        & 60.9 & 87.3 & 68.0 & 52.7 & 61.0 & 65.9 \\
    FitNet \cite{romero2014fitnets} & 60.9 & 87.9 & 68.2 & 52.4 & 61.1 & 67.1 \\
    MasKD \cite{huang2023masked}    & 61.7 & 88.1 & 69.6 & 53.4 & 62.0 & 66.2 \\
    \rowcolor{gray!20}
    LIAF-KD (ours) & 62.7 & 88.7 & 70.9 & 54.9 & 63.1 & 67.6 \\
    \rowcolor{gray!20}
    LIAF-KD+MasKD  & 62.8 & 88.6 & 71.0 & 55.1 & 62.8 & 68.5 \\
    \specialrule{1pt}{0pt}{0pt}
    T: RetinaNet-R101 & 60.1 & 87.4 & 67.1 & 47.7 & 61.6 & 66.4 \\
    S: RetinaNet-R50  & 55.6 & 86.0 & 59.6 & 43.0 & 57.3 & 62.5 \\\hline
    DeFeat \cite{Guo_2021_CVPR}     & 58.3 & 87.2 & 65.5 & 51.0 & 58.4 & 65.9 \\
    FGFI \cite{Wang_2019_CVPR}      & 58.2 & 86.7 & 66.1 & 51.2 & 58.1 & 64.3 \\
    GID \cite{Dai_2021_CVPR}        & 58.2 & 87.1 & 64.8 & 50.0 & 58.3 & 65.7 \\
    FitNet \cite{romero2014fitnets} & 58.1 & 87.2 & 65.1 & 51.3 & 57.8 & 65.7 \\
    MasKD \cite{huang2023masked}    & 58.9 & 87.1 & 65.2 & 45.7 & 60.4 & 65.9 \\
    \rowcolor{gray!20} 
    LIAF-KD (ours) & 59.1 &  86.9 &  65.7 & 46.9 & 60.4 & 67.1 \\
    \rowcolor{gray!20} 
    LIAF-KD+MasKD  & 59.3 & 87.2 & 65.3 & 46.5 & 60.6 & 66.6 \\
    \specialrule{1pt}{0pt}{0pt}
    \end{tabular}}
    \caption{Comparison of object detection KD methods on the KITTI dataset. T: teacher, S: student.}
  \label{tab:kitti-gfl and retinanet}
\end{table}

\noindent\textbf{Results on COCO.}
Table~\ref{tab:coco-gfl and retinanet} exhibits trends similar to those on the KITTI dataset, further confirming the effectiveness of our LIAF-KD on COCO. Interestingly, for the GFL detector, the combined LIAF-KD+MasKD method (42.2\%) performs slightly worse than using our LIAF-KD alone (42.4\%), while still outperforming MasKD (40.6\%). This may be because MasKD considers distracting background pixels, which can conflict with LIAF-KD’s instance-adaptive weighting, introducing misleading guidance during distillation.

\begin{table}[!h]
  \centering
  \resizebox{0.48\textwidth}{!}{
  \begin{tabular}{@{}lcccccc@{}}
    \specialrule{1pt}{0pt}{0pt}
    Method & mAP & $AP_{50}$ & $AP_{75}$ & $AP_{S}$ & $AP_{M}$ & $AP_{L}$ \\\hline
    T: GFL-R101   & 44.9 & 63.1 & 49.0 & 28.0 & 49.1 & 57.2 \\
    S: GFL-R50    & 40.2 & 58.9 & 43.3 & 23.3 & 44.0 & 52.2 \\\hline
    DeFeat \cite{Guo_2021_CVPR}     & 40.8 & 58.6 & 44.2 & 24.3 & 44.6 & 53.7 \\
    FGFI \cite{Wang_2019_CVPR}      & 41.1 & 58.8 & 44.8 & 23.3 & 45.4 & 53.1 \\
    GID \cite{Dai_2021_CVPR}        & 41.5 & 59.6 & 45.2 & 24.3 & 45.7 & 53.6 \\
    FitNet \cite{romero2014fitnets} & 40.7 & 58.6 & 44.0 & 23.7 & 44.4 & 53.2 \\
    MasKD \cite{huang2023masked}    & 40.6 & 58.6 & 43.7 & 22.6 & 43.9 & 53.4 \\
    \rowcolor{gray!20}
    LIAF-KD (ours) & 42.4 & 60.4 & 46.1 & 24.7 & 45.9 & 55.5 \\
    \rowcolor{gray!20}
    LIAF-KD+MasKD  & 42.2 & 60.1 & 45.7 & 24.8 & 45.8 & 55.4 \\
    \specialrule{1pt}{0pt}{0pt}
    T: RetinaNet-R101  & 38.9 & 58.0 & 41.5 & 21.0 & 42.8 & 52.4 \\
    S: RetinaNet-R50   & 37.4 & 56.7 & 39.6 & 20.0 & 40.7 & 49.7 \\\hline
    DeFeat \cite{Guo_2021_CVPR}     & 39.3 & 59.1 & 42.1 & 22.7 & 43.4 & 52.7 \\
    FGFI \cite{Wang_2019_CVPR}      & 38.6 & 58.7 & 41.3 & 21.4 & 42.5 & 51.5 \\
    GID \cite{Dai_2021_CVPR}        & 39.1 & 59.0 & 42.3 & 22.8 & 43.1 & 52.3 \\
    FitNet \cite{romero2014fitnets} & 37.4 & 57.1 & 40.0 & 20.8 & 40.8 & 50.9 \\
    MasKD \cite{huang2023masked}    & 39.8 & 59.0 & 42.5 & 21.5 & 43.9 & 54.0 \\
    \rowcolor{gray!20}
    LIAF-KD (ours) & 39.7 & 58.9 & 42.6 & 21.8 & 43.9 & 53.4 \\
    \rowcolor{gray!20}
    LIAF-KD+MasKD  & 39.9 & 58.9 & 42.7 & 21.8 & 44.2 & 53.9 \\
    \specialrule{1pt}{0pt}{0pt}
    \end{tabular}}
    \caption{Comparison of object detection KD methods on the COCO dataset. T: teacher, S: student.}
  \label{tab:coco-gfl and retinanet}
\end{table}

\subsection{Qualitative Results}

\begin{figure}
  \centering

  
  
  \begin{subfigure}[b]{0.18\linewidth}
    \includegraphics[width=\linewidth]{images/000000049269.jpg}
    \caption{Input}
    \label{fig:sub13}
  \end{subfigure}%
  \hspace{0.1em}
  \begin{subfigure}[b]{0.18\linewidth}
    \includegraphics[width=\linewidth]{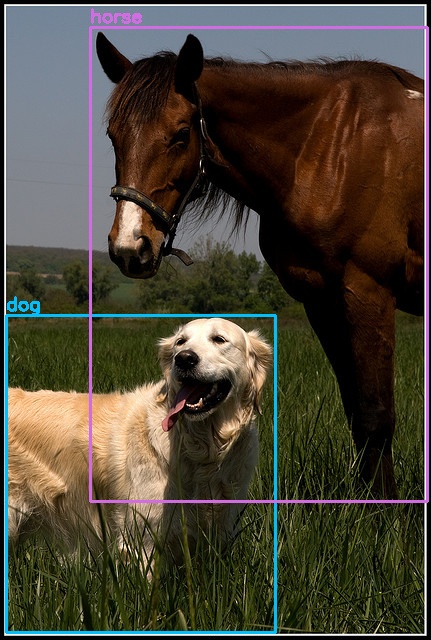}
    \caption{GT}
    \label{fig:sub13-gt}
  \end{subfigure}%
  \hspace{0.1em}
  \begin{subfigure}[b]{0.18\linewidth}
    \includegraphics[width=\linewidth]{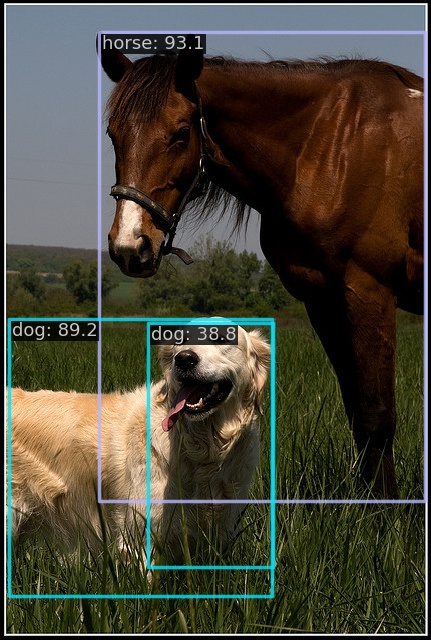}
    \caption{Base}
    \label{fig:sub14}
  \end{subfigure}%
  \hspace{0.1em}
  \begin{subfigure}[b]{0.18\linewidth}
    \includegraphics[width=\linewidth]{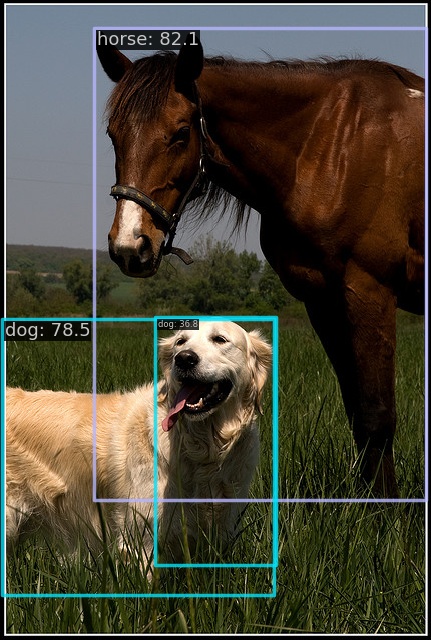}
    \caption{MasKD}
    \label{fig:sub15}
  \end{subfigure}%
  \hspace{0.1em}
  \begin{subfigure}[b]{0.18\linewidth}
    \includegraphics[width=\linewidth]{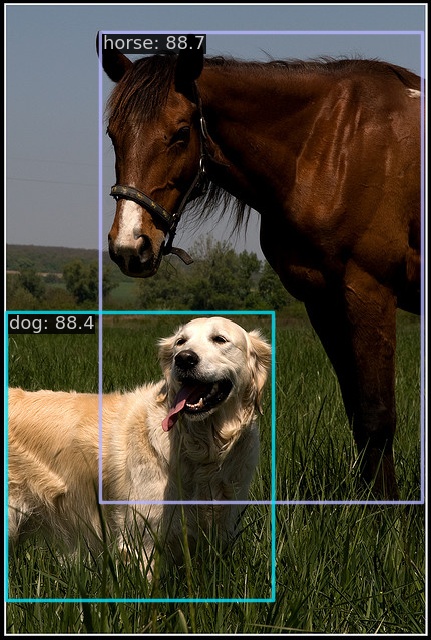}
    \caption{Ours}
    \label{fig:sub16}
  \end{subfigure}
  
  \caption{Detection visualization of different models. Column (a) and (b) are input images and ground truth. Columns (c), (d), and (e) are the detection results of the student baseline, MasKD, and LIAF-KD models, respectively.
  } 
  \label{fig:plotresult} 
\end{figure}

Figure~\ref{fig:plotresult} provides a visual comparison of detection results from the student baseline (c), MasKD (d), and our proposed LIAF-KD (e), with columns (a) and (b) showing the original image and its ground truth bounding boxes. As illustrated, both the student baseline and MasKD often produce incomplete or inaccurate bounding boxes. In contrast, LIAF-KD achieves more accurate object localization and classification.

\begin{figure}
  \centering

  

  \begin{subfigure}[b]{0.18\linewidth}
    \includegraphics[width=\linewidth]{images/figures/000000005529.jpg}
    \caption{Input}
    \label{fig:sub13}
  \end{subfigure}%
  \hspace{0.1em}
  \begin{subfigure}[b]{0.18\linewidth}
    \includegraphics[width=\linewidth]{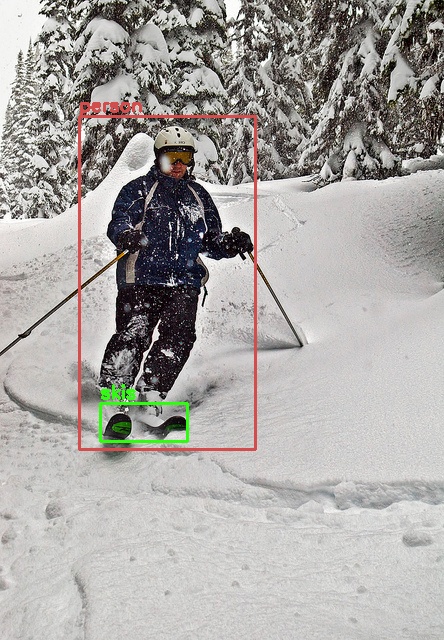}
    \caption{GT}
    \label{fig:sub13-gt}
  \end{subfigure}%
  \hspace{0.1em}
  \begin{subfigure}[b]{0.18\linewidth}
    \includegraphics[width=\linewidth]{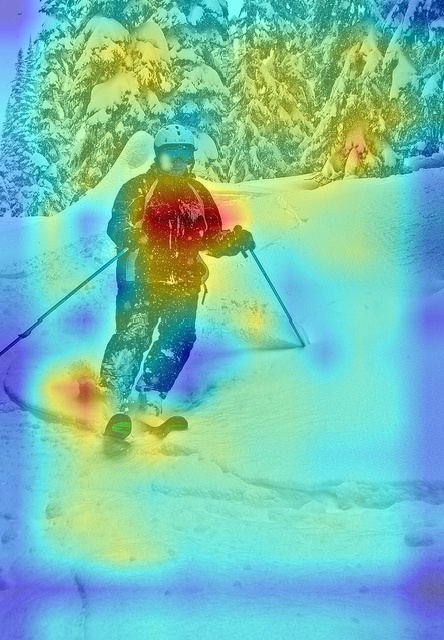}
    \caption{Base}
    \label{fig:sub14}
  \end{subfigure}%
  \hspace{0.1em}
  \begin{subfigure}[b]{0.18\linewidth}
    \includegraphics[width=\linewidth]{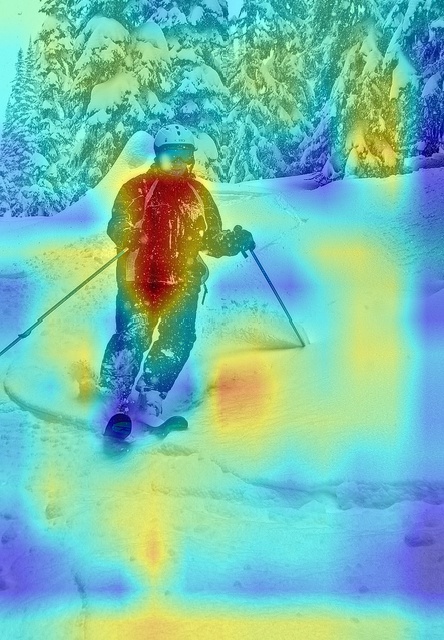}
    \caption{MasKD}
    \label{fig:sub15}
  \end{subfigure}%
  \hspace{0.1em}
  \begin{subfigure}[b]{0.18\linewidth}
    \includegraphics[width=\linewidth]{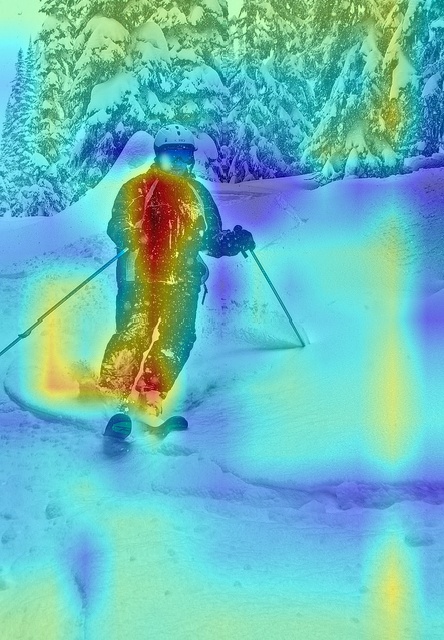}
    \caption{Ours}
    \label{fig:sub16}
  \end{subfigure}

    \caption{Grad-CAM Attention of different models. The colors represent attention intensity, with red indicating the highest level and blue the lowest. ``Base": the student baseline.} 
  \label{fig:attention} 
\end{figure}

Figure~\ref{fig:attention} compares the attention maps generated by different models. Without student-aware instance selection, MasKD and the student baseline disperse attention across irrelevant background areas and under-emphasize critical instance regions. On the other hand, LIAF-KD yields attention maps that are more focused and precisely aligned with meaningful object regions.

\subsection{Efficiency Analysis}

Table~\ref{tab:my_awesome_table_v2} presents the efficiency improvements achieved by our method on different detectors (top: RetinaNet; bottom: GFL), while maintaining comparable accuracy.

\begin{table}[!h]
\centering
\resizebox{0.48\textwidth}{!}{
\begin{tabular}{lccccc}
\toprule
Method & mAP & Params(M) & FLOPs(T) & FPS(GPU) & FPS(CPU)\\
\midrule
T:Retina-R101 & 38.9 & 57.0 & 0.29 & 33.5 & 0.935\\ 
S: LIAF-KD    & 39.7 & 38.0 & 0.22 & 44.5 & 1.308 \\\hline
T: GFL-R101   & 44.9 & 51.4 & 0.26 & 30.3 & 1.106 \\
S: LIAF-KD    & 42.4 & 32.4 & 0.19 & 40.0 & 1.335 \\
\bottomrule
\end{tabular}}
\captionof{table}{Efficiency comparison of different object detectors. We report inference speed on an NVIDIA A100 GPU and an AMD EPYC™ 7H12 CPU. T: teacher, S: student.}
\label{tab:my_awesome_table_v2}
\end{table}
\section{Conclusion} \label{sec:conclusion}

This paper presented Learnable Instance Attention Filtering for Adaptive Detector Distillation (LIAF-KD), a framework designed to address the limitations of existing feature-based KD methods that rely on heuristic or teacher-prescribed attention. Unlike prior instance-agnostic approaches,
LIAF-KD introduces learnable instance selectors that evaluate the relative importance of object instances based on their feature representations, enabling a more adaptive and informative distillation process that is also aware of the student’s evolving learning state. Extensive experiments on the KITTI and COCO datasets demonstrate the effectiveness of LIAF-KD across diverse detectors. The proposed method consistently outperforms student baselines and several state-of-the-art KD approaches while maintaining a favorable complexity–accuracy trade-off, highlighting its potential for real-time and resource-constrained applications.

\bibliographystyle{IEEEtran}
\bibliography{refs}

\end{document}